\documentclass{article}

\PassOptionsToPackage{numbers, compress}{natbib}
\usepackage[preprint]{nips_2018}

\usepackage[utf8]{inputenc} % allow utf-8 input
\usepackage[T1]{fontenc}    % use 8-bit T1 fonts
\usepackage{hyperref}       % hyperlinks
\usepackage{url}            % simple URL typesetting
\usepackage{booktabs}       % professional-quality tables
\usepackage{amsfonts}       % blackboard math symbols
\usepackage{nicefrac}       % compact symbols for 1/2, etc.
\usepackage{microtype}      % microtypography
\usepackage{bm}
\usepackage{amsmath}
\usepackage{amssymb}

\usepackage{graphicx}
\usepackage[hang,centerlast,nooneline]{subfigure}

\usepackage{algorithm}
\usepackage[noend]{algorithmic}

\DeclareMathOperator*{\argmin}{argmin}

\DeclareMathOperator*{\conv}{conv}

\usepackage[most]{tcolorbox}
\tcbset{
    frame code={}
    center title,
    left=10pt,
    right=0pt,
    top=2pt,
    bottom=3pt,
    colframe=white,
    width=\linewidth,
    leftrule=-10pt,
    rightrule=0pt,
    toprule=0pt,
    bottomrule=-2pt,
    boxsep=0pt,
    arc=0pt,outer arc=0pt,
    }

\usepackage{xcolor}

\definecolor{green_fw}{RGB}{0,150,136}
\definecolor{lightgreen_fw}{RGB}{139,195,74}
\definecolor{cyan_fw}{RGB}{0,188,212}
\newtcolorbox{purplebox}{colback=red!40!blue!35!}
\newtcolorbox{greenbox}{colback=white!30!green_fw}
\newtcolorbox{lightgreenbox}{colback=white!40!lightgreen_fw}
\newtcolorbox{cyanbox}{colback=white!20!cyan_fw}

\title{Neural Conditional Gradients}

\author{
  Patrick Schramowski%\thanks{Use footnote for providing further
    \\
  Department of Computer Science\\
  TU Darmstadt, Germany\\
  \texttt{schramowski@cs.tu-darmstadt.de} \\
  \And
  Christian Bauckhage\thanks{Bonn-Aachen International Center for Information Technology, Univ.~of Bonn, Germany} \\
  Fraunhofer IAIS \\ 
  Bonn, Germany \\
  \texttt{christian.bauckhage@iais.fraunhofer.de} \\
  \AND
  Kristian Kersting\thanks{Center for Cognitive Science, TU Darmstadt, Germany} \\
  Department of Computer Science \\
  TU Darmstadt, Germany \\
  \texttt{kersting@cs.tu-darmstadt.de} \\
}

\begin{document}

\maketitle

\begin{abstract}
  The move from hand-designed to learned optimizers in machine learning has been quite successful for gradient-based and -free optimizers. When facing a constrained problem, however, maintaining feasibility typically requires a projection step, which might be computationally expensive and not differentiable. We show how the design of projection-free convex optimization algorithms can be cast as a learning problem based on Frank-Wolfe Networks: recurrent networks implementing the Frank-Wolfe algorithm aka.~conditional gradients. This allows them to learn to exploit structure when, e.g., optimizing over rank-1 matrices. Our LSTM-learned optimizers outperform hand-designed as well learned but unconstrained ones. We
demonstrate this for training support vector machines and softmax classifiers.
\end{abstract}

\section{Introduction}
\label{introduction}
\begin{figure}[t!]
\begin{minipage}[t]{\dimexpr.5\textwidth-0.5em}
\hrule\vspace{1mm}
\textbf{Frank-Wolfe Network (FWNet)} 
\vspace{1mm}\hrule
\begin{center}
		\includegraphics[width=0.805\columnwidth]{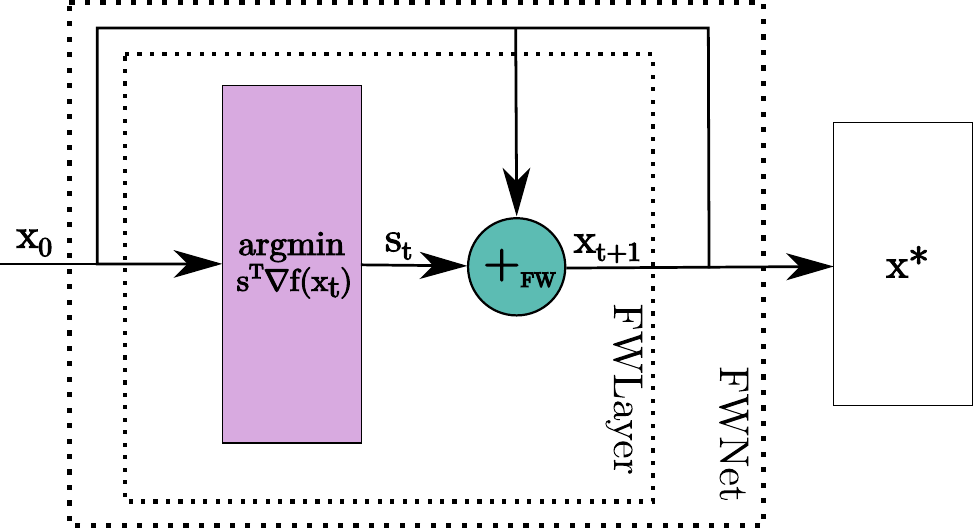}\\
      	\vspace{1mm}
		\hrule\vspace{1mm}
\end{center}
\textbf{Frank-Wolfe (FW) algorithm}~(\citeyear{fw_frank_wolfe})\vspace{2mm}\label{alg:fw}
\vspace{1mm}\hrule
\begin{algorithmic}[1]
\STATE Let $\mathbf{x}_0\in\pmb{S}$
\FOR{$t=1,2,3,\ldots,T$}
\STATE Choose step-size $\gamma_t \in [0,1]$, \\ e.g., $\gamma_t =\frac{2}{t+1}$
\begin{purplebox}
\STATE Compute $\mathbf{s}_t =\argmin\nolimits_{{s \in \pmb{S}}} \ \mathbf{s}^T \nabla f(\mathbf{x}_t)$
\end{purplebox}
\begin{greenbox}
\STATE Update $\mathbf{x}_{t+1}=(1-\gamma_t)\mathbf{x}_{t} + \gamma_t\mathbf{s}_t$
\end{greenbox}
\ENDFOR
\end{algorithmic}
\hrule
		\caption{FWNets (top) implement the Frank-Wolfe algorithm (bottom) as recurrent neural networks. Unrolled over time, a FWNet layer takes the current state $\pmb{x}_{t}$ as input, computes the linearization (purple layer) and moves the next internal state $\pmb{x}_{t+1}$ towards a convex minimizer of $\pmb{x}_{t}$ and this linearization (green layer). \label{fig:networks_fw_general}}
%	\end{center}
\vskip -0.2in
\end{minipage}\hfill
\begin{minipage}[t]{\dimexpr.5\textwidth-0.5em}
\hrule\vspace{1mm}
\textbf{Neural Support Vector Machine} 
\vspace{1mm}\hrule %\vspace{1mm}
\begin{center}
		\includegraphics[width=1.0\columnwidth]{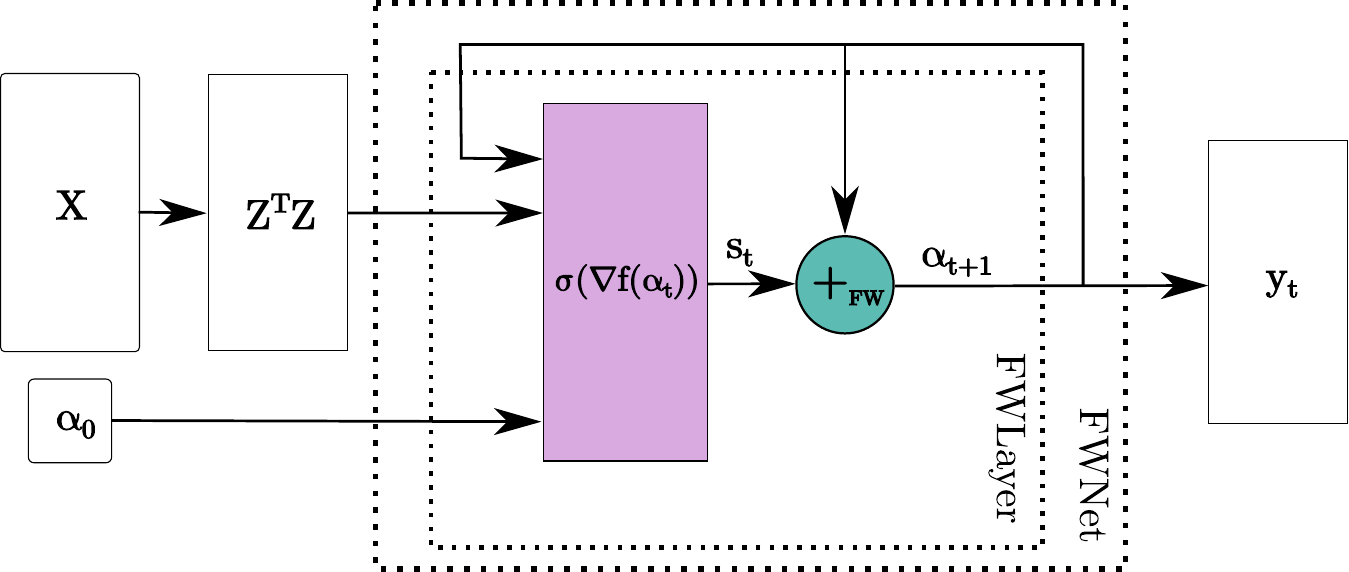}\\
%      	\vspace{-2mm}
        \end{center}
		\hrule\vspace{1mm}
\textbf{Frank-Wolfe (FW) algorithm over the unit simplex}\label{alg:fwsvm}
\vspace{1mm}\hrule
\begin{algorithmic}[1]
\STATE Let $\mathbf{\alpha}_0\in\pmb{S}$
\FOR{$t=1,2,3,\ldots,T$}
\STATE Choose step-size $\gamma_t \in [0,1]$, \\ e.g., $\gamma_t =\frac{2}{t+1}$
\begin{purplebox}
\STATE Compute $\mathbf{s}_t =\pmb{\sigma}(\nabla f(\pmb{\alpha}_t))$\vspace{0.45mm}
\end{purplebox}
\begin{greenbox}
\STATE Update $\pmb{\alpha}_{t+1}=(1-\gamma_t)\pmb{\alpha}_{t} + \gamma_t\mathbf{s}_t$ 
\end{greenbox}
\ENDFOR
\end{algorithmic}
\hrule
		\caption{A FWNet (top) implementing Support Vector Machines (SVMs). The gram matrix $Z^TZ$ is treated as synaptic connections of $n$ neurons and is kept fixed over time. Each unrolled FWNet layer takes it and the support vector coding $\pmb{\alpha}_{t}$ as input, computes $\sigma(\nabla f(\pmb{\alpha}_t))$ as linearization (purple layer) and moves the next support vector coding $\pmb{\alpha}_{t+1}$ towards a convex minimizer of $\pmb{\alpha}_{t}$ and this linearization (dark green layer). This is a neural instantiation of the FW algorithm for optimization over the unit simplex (bottom). \label{fig:networks_fw_bauckhage}}
%	\end{center}
\vskip -0.2in
$\ $ % do not remove this
\end{minipage}
\end{figure}

Machine learning tasks can often be expressed as general constrained convex optimization problems of the form
\begin{equation}
\pmb{x^*} = \argmin\nolimits_{\pmb{x} \in \pmb{S}} \ f(\pmb{x})\;,
\end{equation}
where $f$ % :
is a convex and continuously differentiable function, and $\pmb{S}$ 
is a compact convex subset of a Hilbert space. For such
optimization problems, one of the simplest and earliest
known iterative optimizers is given by the Frank-Wolfe (FW) algorithm~\cite{fw_jaggi13}, summarized in Fig.~\ref{fig:networks_fw_general}(bottom), also known as \textit{conditional gradient}. In each iteration, it considers
the linearization of the objective at the current position $\mathbf{x}$ and moves
towards a convex minimizer of this linear function (taken
over the same domain). In other words, Frank-Wolfe effectively turns the constrained convex optimization problem 
into a series of simple linear optimization problems.
%\begin{figure}[t]
%\hrule\vspace{1mm}
%\textbf{Frank-Wolfe Network (FWNet)} 
%\vspace{1mm}\hrule
%\begin{center}
%		\includegraphics[width=0.6\columnwidth]{images/networks/fw_net_general}\\
%      	\vspace{1mm}
%		\hrule\vspace{1mm}
%\end{center}
%\textbf{Frank-Wolfe (FW) algorithm}~(\citeyear{fw_frank_wolfe})\label{alg:fw}
%\vspace{1mm}\hrule
%\begin{algorithmic}[1]
%\STATE Let $\mathbf{x}_0\in\pmb{S}$
%\FOR{$t=1,2,3,\ldots,T$}
%\STATE Choose step-size $\gamma_t \in [0,1]$, e.g., $\gamma_t =\frac{2}{t+1}$
%\begin{purplebox}
%\STATE Compute $\mathbf{s}_t =\argmin\nolimits_{{s \in \pmb{S}}} \ \mathbf{s}^T \nabla f(\mathbf{x}_t)$
%\end{purplebox}
%\begin{greenbox}
%\STATE Update $\mathbf{x}_{t+1}=(1-\gamma_t)\mathbf{x}_{t} + \gamma_t\mathbf{s}_t$
%\end{greenbox}
%\ENDFOR
%\end{algorithmic}
%\hrule
%		\caption{FWNets (top) implement the Frank-Wolfe algorithm (bottom) as recurrent neural networks. Unrolled over time, a FWNet layer takes the current state $\pmb{x}_{t}$ as input, computes the linearization (purple layer) and moves the next internal state $\pmb{x}_{t+1}$ towards a convex minimizer of $\pmb{x}_{t}$ and this linearization (green layer). \label{fig:networks_fw_general}}
%	\end{center}
%\vskip -0.2in
%\end{figure}

Recently, Bauckhage (\citeyear{fw_neural}) employed this view to implement Frank-Wolfe optimizing over the unit simplex---the convex hull $\pmb{S}:=\conv(\{ \mathbf{e}_i|i\in [n ]\})$ of the unit
basis vectors---in terms of a recurrent neural network (RNN). Since the domain $\pmb{S}$ is 
given as an intersection
of linear constraints, the subproblems can be solved using softmin 
activation functions. 
This paper significantly extends our understanding of such neural conditional gradients. 

As warm up, we show that the resulting Frank-Wolfe Networks (FWNets)---the generalized architecture is shown in Fig.~\ref{fig:networks_fw_general}(top)---allow one to implement (training) support vector machines 
directly within neural networks. Unfortunately, the resulting neural optimizer is too dense to scale to large classification problems: it hinges on the quadratic gram matrix. 
Consequently, as our second contribution, we introduce sparse FWNets for convex optimization over the unit ball of the \textit{trace-norm}, i.e., $\pmb{S}:=\conv(\{ \mathbf{u}\mathbf{v}^T|\mathbf{u}\in\mathbb{R}^n, ||\mathbf{u}||_2=1 \ \text{and} \ \mathbf{v}\in\mathbb{R}^n, ||\mathbf{v}||_2=1\})$. Since the subproblems amount to approximating the unit left and right top singular vectors of the gradient
matrix $\nabla f(\mathbf{w}_t)$, we replace the softmin activation functions by sparse RNNs that are structurally equivalent to the well known power iteration. This allows one to realize neural conditional gradients for, e.g., sparse softmax classifiers that scale well to large datasets. 

The closest in spirit to FWNets are probably OptNets~\cite{amos17}. They 
integrate constrained optimization problems, in particular quadratic ones as individual layers into neural networks.
This has also the potential of richer end-to-end training for complex tasks that require such optimization. However, OptNets do not cast the optimizer itself as a neural network. Instead, external optimizers are invoked to solve OptNets. 
This hampers a seamless integration with other deep learning concepts. 

Consider e.g.~{\it learning to learn} (L2L), which has a long history in psychology \cite{ward37,harlow49,kehoe88} and has inspired many recent 
attempts within the machine learning community to build agents capable of learning to learn \cite{schmidhuber87,naik92,thrun98,hochreiter01,santoro16,l2l,wang16,ravi17,li17}. So far, however, learning to learn has mainly been considered for gradient(-free) optimizers (L2LG); \textit{learning to learn by conditional gradients} (L2LC) has not been proposed. Our third contribution fills this gap. We show how to
boost the performance of neural conditional gradients by learning parts of them instead of using hand-coded ones. Our learned conditional gradient optimizers, implemented by LSTMs, outperform hand-designed as well as unconstrained but learned competitors. We
demonstrate this on a number of classification tasks, including training deep SVM and softmax classifiers. 

We proceed as follows. We start of by reviewing L2L. Then we illustrate FWNets and use them to devise L2LC. Afterwards we introduce sparse FWNets for \textit{trace-norm} problems. Before concluding, we present our experimental evaluation. 

\section{Learning to learn by gradients by gradients}
Let us start off by briefly reviewing learning to learn by gradients by gradients \cite{l2l}. The
goal is to optimize an objective function $f(\theta)$ defined over some domain $\theta \in \Theta$. 
To this end, we find the minimizer $\theta^* = \argmin_{\theta\in\Theta} f(\theta)$. While any method capable of minimizing this objective can be applied, the standard approach for differentiable functions is some form of gradient descent, resulting in a sequence of updates
%\begin{equation}
$\theta_{t+1} =\theta_t - \alpha_t \nabla(\theta_t).$
%\end{equation}
To realize L2L, \citeauthor{l2l}~(\citeyear{l2l}) proposed to replace hand-designed update rules with a learned update rule, called the optimizer $g$, specified by its own set of parameters. 
This results in updates to the optimizee $f$ of the form 
%\begin{equation}
$\theta_{t+1} =\theta_{t} + g_t(\nabla(\theta_t), \phi).$
%\end{equation}
More precisely, \citeauthor{l2l} advocated to realize the update rule $g$ using a recurrent neural network (RNN), which maintains its own state and hence dynamically updates as a function of its iterates.
%\label{related_work}

Indeed this learning to learn by gradients by gradients is widely applicable due to the simplicity of gradient computations. When facing a constraint optimization problem, however, maintaining feasibility typically requires a projection step, which is potentially computationally expensive, especially for complex feasible regions in very large dimensions. 
To overcome this, we advocate the use of the Frank-Wolfe algorithm~\cite{fw_jaggi13}, which
eschews the projection step and rather use a linear optimization oracle to stay within the feasible region. 
While convergence rates and regret bounds are often suboptimal, in many cases the gain due to only having 
to solve a single linear optimization problem over the feasible region in every iteration still 
leads to significant computational advantages. This may explain its popularity for problems such as computing the distance to a convex hull, computing a minimum enclosing ball, or training a support vector machine. 

\section{Neural support vector machines}
\label{sec:fw_svm}
Support Vector Machine (SVM) are working horses of machine learning.
Frank Wolfe algorithms for training them~\cite{ouyangG10} 
solve a quadratic program (QP) over the unit simplex, i.e., the convex hull $\pmb{S}:=\conv(\{ \mathbf{e}_i|i\in [n ]\})$ of the unit
basis vectors. 
To implement them as neural networks, we can proceed as follows. 
Consider, e.g., the $l_2$-SVM formulation for binary classification
\begin{equation}
\min \left( \frac{1}{2} w^2 - p + \frac{C}{2} \sum\nolimits_{i=1}^{N} \epsilon_i^2\right)  \  s.t. \  \mathbf{w}^T \mathbf{z}_i \geq p - \xi_i \nonumber
\end{equation}
where $\mathbf{z}_i = y_i \mathbf{x}_i$. The corresponding Lagrangian dual problem for SVMs can be expressed as:
\begin{equation}
\label{eq:dual_svm}
\min\nolimits_{\pmb{\alpha} \in \mathbb{R}^m} f(\pmb{\alpha}) = \frac{1}{2}\pmb{\alpha}^T\pmb{K} \pmb{\alpha}, \alpha_i \geq 0, \sum\nolimits_{i=1}^{N}\alpha_i = 1. 
\end{equation}
Here, $\pmb{K}$ is a positive definite kernel matrix 
%\begin{equation}
 $\pmb{K} = \pmb{Z}^T\pmb{Z} = (\pmb{y} \circ \pmb{x})^T(\pmb{y} \circ \pmb{x})\;.$ % \nonumber
%\end{equation}
As shown in \cite{fw_svm2}, $H(\pmb{\Sigma}) = \{\pmb{e}_1, \dots, \pmb{e}_n\}$, hence we have $\pmb{s}_t = \pmb{e}^{i^*}_t$ where
\begin{eqnarray}
i^*_t &\in& \argmin\nolimits_{i=1, \dots, m} \  \nabla f(\alpha_t)^{(i)} \nonumber \\
%\end{eqnarray}
%\begin{eqnarray}
&=& \argmin\nolimits_{i=1, \dots, m} \sum\nolimits_{j | \alpha^{(j)}_t > 0 } K^{(i,j)}\alpha^{(j)}_{t}. \nonumber
\end{eqnarray}
So, the gradient for the new objective is $\nabla f(\pmb{\alpha}_t) = \bm{K} \bm{\alpha}$. Therefore FW requires computing
\begin{equation}
\bm{s}_t = \argmin\nolimits_i \bm{e}_i^T(\bm{K}\bm{\alpha})\approx\pmb{\sigma}(\bm{K}\bm{\alpha})
\label{softmin_fw}
\end{equation}
where the non-linear, vector-valued function $\pmb{\sigma}(\pmb{z})$ is the well known softmin
operator defined as
\begin{equation}
\pmb{\sigma}(\pmb{z})_i = \frac{\exp(-\beta z_i)}{\sum\nolimits_j \exp(-\beta z_j)}
\end{equation}
for which we note %that
%\begin{equation}
$\lim\nolimits_{\beta \rightarrow \infty} \pmb{\sigma}(z) = \pmb{e}_i = \argmin\nolimits_j \  \pmb{e}_j^T\pmb{z}\;.$
%\end{equation}
Plugging in the relaxed optimization step~\eqref{softmin_fw}, we can now rewrite the Frank-Wolfe updates for SVMs as
\begin{eqnarray}
\pmb{\alpha}_{t+1} &=& \pmb{\alpha}_t  + \gamma_t\pmb{d}_t
= \pmb{\alpha}_t  + \gamma_t(\pmb{s}_t - \pmb{\alpha}_t) \nonumber \\
&=& (1- \gamma_t)\pmb{\alpha}_t + \gamma_t\pmb{s}_t %\nonumber \\
\approx (1-\gamma_t)\pmb{\alpha}_t + \gamma_t\pmb{\sigma}(\pmb{K}\pmb{\alpha}_t)\;,\nonumber 
\end{eqnarray}
where $ \pmb{K}\pmb{\alpha}_t=\nabla f(\pmb{\alpha})$.
But this is then to say that by choosing an appropriate parameter for the softmin function the following non-linear dynamical system
\begin{eqnarray}
\pmb{\alpha}_{t+1} = (1-\gamma_t)\pmb{\alpha}_t + \gamma_t\pmb{\sigma}(\nabla f(\pmb{\alpha}_t)),
\label{eq:rnn_svm}
\end{eqnarray}
mimics Frank-Wolfe up to arbitrary precision. 

%\begin{figure}[t]
%\hrule\vspace{1mm}
%\textbf{Neural Support Vector Machine} 
%\vspace{1mm}\hrule %\vspace{1mm}
%\begin{center}
%		\includegraphics[width=0.65\columnwidth]{images/networks/fw_net_svm}\\
%%      	\vspace{-2mm}
%        \end{center}
%		\hrule\vspace{1mm}
%\textbf{Frank-Wolfe (FW) algorithm over the unit simplex}\label{alg:fwsvm}
%\vspace{1mm}\hrule
%\begin{algorithmic}[1]
%\STATE Let $\mathbf{w}_0\in\pmb{S}$
%\FOR{$t=1,2,3,\ldots,T$}
%\STATE Choose step-size $\gamma_t \in [0,1]$, e.g., $\gamma_t =\frac{2}{t+1}$
%\begin{purplebox}
%\STATE Compute $\mathbf{s}_t =\pmb{\sigma}(\nabla f(\pmb{\alpha}_t))$
%\end{purplebox}
%\begin{greenbox}
%\STATE Update $\pmb{\alpha}_{t+1}=(1-\gamma_t)\pmb{\alpha}_{t} + \gamma_t\mathbf{s}_t$ 
%\end{greenbox}
%\ENDFOR
%\end{algorithmic}
%\hrule
%		\caption{A FWNet (top) implementing Support Vector Machines (SVMs). The gram matrix $Z^TZ$ is treated as synaptic connections of $n$ neurons and is kept fixed over time. Each unrolled FWNet layer takes it and the support vector coding $\pmb{\alpha}_{t}$ as input, computes $\sigma(\nabla f(\pmb{\alpha}_t))$ as linearization (purple layer) and moves the next support vector coding $\pmb{\alpha}_{t+1}$ towards a convex minimizer of $\pmb{\alpha}_{t}$ and this linearization (dark green layer). This is a neural instantiation of the FW algorithm for optimization over the unit simplex (bottom). \label{fig:networks_fw_bauckhage}}
%%	\end{center}
%\vskip -0.2in
%\end{figure}

The underlying FW over the unit simplex is summarized in Fig.~\ref{fig:networks_fw_bauckhage}(bottom). Structurally it is 
equivalent to the system of equations 
governing the dynamics of echo state networks,
a particular from of recurrent neural networks (RNN), shown in Fig.~\ref{fig:networks_fw_bauckhage}(top) for training SVMs. For inference, 
we can unroll the RNN into a multi-layer neural network. Due to well known 
FW convergence results~\cite{fw_jaggi13}, we know that $\mathcal{O}(1\slash\epsilon)$ 
layers are likely to provide an $\epsilon$-approximate solution to the SVM problem. 
\begin{figure}[t]
	\begin{center}
		\includegraphics[width=0.45\columnwidth]{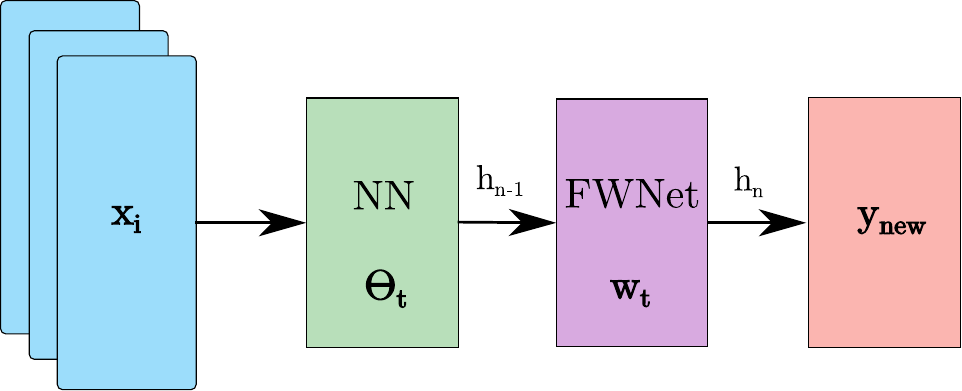}
		\caption{Deep SVMs: Stacking Frank-Wolfe Networks (FWNets) for SVMs on top of a deep neural network.\label{fig:networks_nn+fw}}
	\end{center}
\end{figure}
\begin{figure}[b]
\begin{center}
\includegraphics[width=0.6\columnwidth]{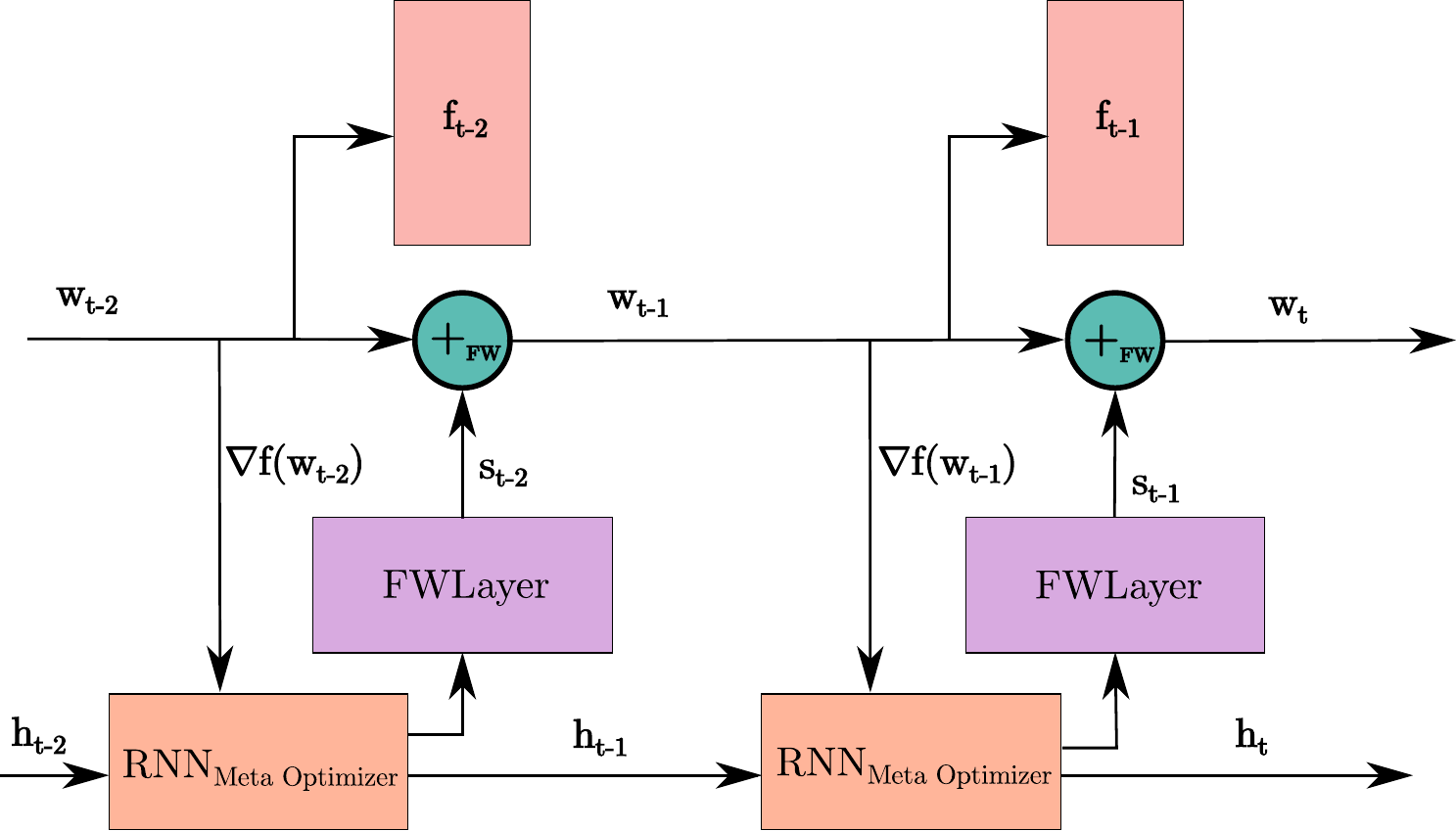}
\caption{Learning to learn by conditional gradients by gradients. The optimizer, an FWNet, is unrolled over time, resulting in FWLayers (purple boxes). They are trained by an RNN, the optimizee (orange boxes).\label{fig:networks_l2lc}}
\end{center}
\end{figure}
Moreover, the neural view on training SVMs allows one to deepify SVMs:
we replace the final classification layer of a deep network by a FWNet that trains an SVM. This enables end-to-end training akin to \cite{deepsvm2,deepsvm1} but in a simpler and fully neural fashion:
the SVM parameters are updated via a forward-propagation only, and the parameters of the kernel neural network are updated by gradient descent using back-propagation of the error starting at the FWNet, cf.~Fig.~\ref{fig:networks_nn+fw}. Here, $\pmb{h}_{n-1}$ denotes the input to the FWNets. During training, the FWNet computes the Kernel $\pmb{K}$ at each iteration and the weights $\pmb{w}$ (respectively $\pmb{\alpha}$ of Eq.~\eqref{eq:rnn_svm}) are updated as described above.
To predict the class of a new example $\pmb{x}_{new}$, we make one foward-pass through the network. 

\begin{figure*}[t!]
\begin{minipage}[t]{\dimexpr.5\textwidth-0.5em}
\hrule \ \\
\textbf{Frank-Wolfe algorithm for optimization over low-rank matrices using power iterations}\label{alg:fw_trace}
\vspace{1mm}\hrule
\begin{algorithmic}[1]
\STATE Let $\pmb{w}_0 \sim \mathcal{N}(0, 1)$
\FOR{$t=0,1,\ldots,T-1$}
\STATE Choose step-size $\gamma_t \in [0,1]$,\\ e.g., $\gamma_t =\frac{2}{t+2}$
\begin{lightgreenbox}
\STATE Compute $\nabla F(\pmb{w}_t)$
\end{lightgreenbox}
\begin{purplebox}
\STATE Set $\pmb{v}_0 \in R^m$ uniformly from unit sphere
\FOR{$k=0,1,\ldots,K-1$}
\STATE $\pmb{u}_{k+1} \leftarrow \nabla F(\pmb{w}_t)\pmb{v}_k$
\STATE $\pmb{u}_{k+1} \leftarrow \pmb{u}_{k+1}/||\pmb{u}_{k+1}||$
\STATE $\pmb{v}_{k+1} \leftarrow \nabla F(\pmb{w}_t)^T\pmb{u}_k$ 
\STATE $\pmb{v}_{k+1} \leftarrow \pmb{v}_{k+1}/||\pmb{v}_{k+1}||$
\ENDFOR
\end{purplebox}
\begin{cyanbox}
\STATE $\pmb{s}_t = -\mu \pmb{u}_1\pmb{v}_1^T$
\end{cyanbox}
\begin{greenbox}
\STATE Update $\pmb{w}_{t+1}=(1-\gamma_t)\pmb{w}_{t} + \gamma_t\pmb{s}_t$
\end{greenbox}
\ENDFOR
\end{algorithmic}
\vspace{3ex}
\hrule 
\end{minipage}\hfill
\begin{minipage}[t]{\dimexpr.5\textwidth-0.5em}
\hrule \ \\
\textbf{FWNets for optimization over low-rank matrices using neural power iterations} 
\vspace{1mm}\hrule
\begin{center}
		\includegraphics[width=\textwidth]{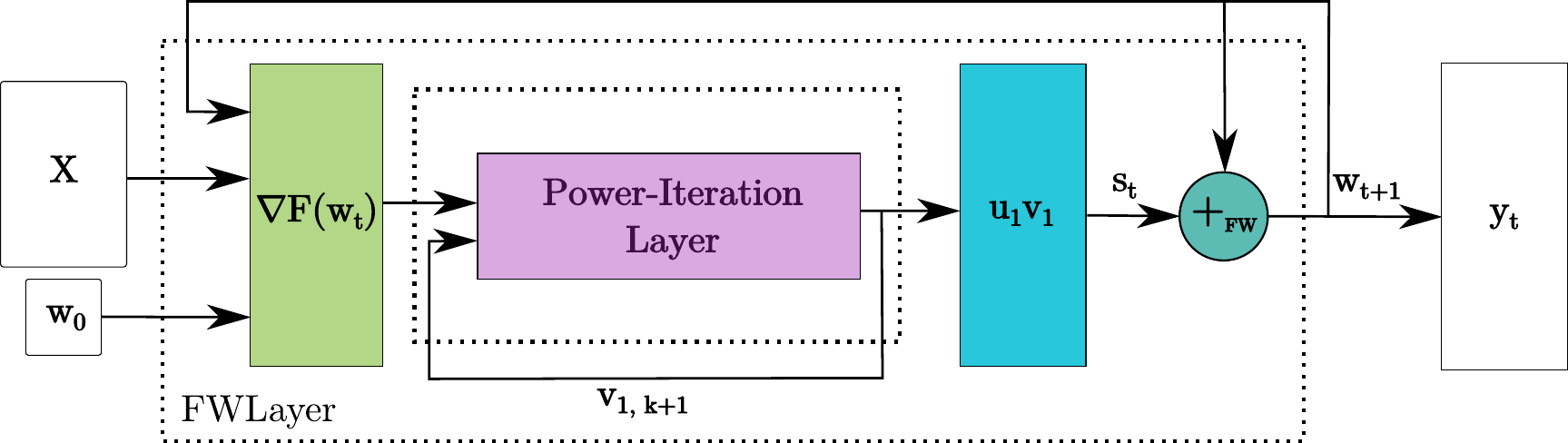}\\ \vspace{4mm} 
        \includegraphics[width=0.759\textwidth]{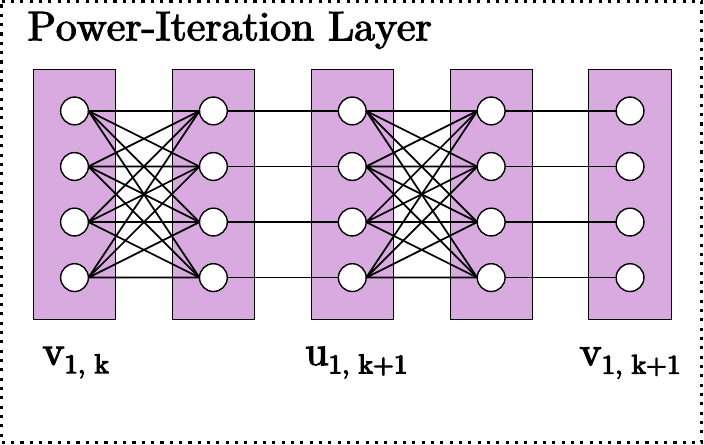}
       \end{center}
       \vspace{1.1mm}
        \hrule
\end{minipage}
		\caption{On bounded trace-norm domain, the subproblems of FW(left) amount to approximating the unit left and right top singular vectors of the gradient
matrix $\nabla F(\mathbf{w}_t)$. To implement this within FWNets (right), RNNs implementing power iterations are used.\label{fig:networks_fw_trace_with_power_method}}
\end{figure*}

\section{Learning to learn by conditional gradients by gradients (L2LC)}
\label{l2cl}
The performance of FWNets is hampered by the fact that they only makes use of the linearizations, ignoring other information such as curvature. To speed them up, we now introduce {\it learning to learn by conditional gradients by gradients} (L2LC)
as depicted in Fig.~\ref{fig:networks_l2lc}.

The FWNet is unrolled over time $t$, and at each step $t$ parts of the optimizer are trained using an RNN as optimize (orange). This way, the optimizee adapts the parts, which are then used to form a Frank-Wolfe update. Consider, e.g., learning to learn SVMs. Instead of using the typical hand-coded rule $\gamma = 2/(2+t)$ or implementing a line-search, we learn to adapt, e.g., the learning-rate:
\begin{eqnarray}
\label{eq:l2cl}
L(\pmb{\gamma}) &=&  \mathbb{E}_f \left[\sum\nolimits^{T}_{t=1}f(\pmb{w}_t)\right] \text{where} \nonumber \\
\gamma_{t}, \pmb{h}_{t+1} &=& RNN(\gamma_{t-1}, \pmb{h}_t, \phi) \\
\pmb{w}_{t+1} &=& (1-\gamma_t)\pmb{w}_t + \gamma_t\pmb{\sigma}(\nabla f(\pmb{w}_t)) \nonumber \
\end{eqnarray}
where $\pmb{w}$ are the weights of the Frank-Wolfe layer and \pmb{h} the state of an RNN, e.g., an LSTM. Or, we learn to adapt the conditional gradient itself. For that, one has to be little bit more careful. We have to ensure that the predictions are on the unit simplex:
\begin{eqnarray}
g_t, \pmb{h}_{t+1} &=& RNN(\nabla f(\pmb{w}_t), \pmb{h}_t, \phi)\\
\pmb{w}_{t+1} &=& (1-\gamma_t)\pmb{w}_t + \gamma_t\pmb{\sigma}(g_t) \nonumber \
\end{eqnarray}
where $\gamma_t=2/(t+2)$ or $\gamma_t$ is constant. 
That is, the RNN predicts the unconstrained gradient, which is then projected onto the unit simplex using a sigmoid. Overall, the FWNet is unrolled over the learning iterations $t$, and at each step $t$ the unconstrained gradient $\nabla f(w_t)$ is used as input to the RNN, the optimizee (orange). The prediction in then squeezed through a sigmoid and we update the weight vector. 

\section{Neural sparse softmax classifiers}
\label{fw_optimizing_low-rank_matrices}
Unfortunately, neural SVMs are not likely to scale well. The underlying SVM scales quadratically in the number of training examples due to the gram matrix. Indeed, one may resort to devise neural implementation of stochastic Frank-Wolfe algorithms~\cite{fw_jaggi13} or frame the learning problem within L2LC, generalizing local FWNets to a global model~\cite{vinyalsBLKW16,ravi17}. Here we introduce FWNets for training large-scale, sparse softmax classifiers~\cite{fw_multi_class_classification}, i.e., for optimization problems of the following form:
\begin{eqnarray}\label{eq:fw_multiclass_objective_function}
 \min\nolimits_{\pmb{w} \in \pmb{M}} F(\pmb{w}) = \frac{1}{n} \sum\nolimits_{i=1}^{n} f_i(\pmb{w})
\end{eqnarray}
where
$\pmb{M} = \{w \in \mathbb{R}^{h \times m} | \ ||\pmb{w}||_* \leq \tau\}$ with $||\cdot||_*$ being the trace-norm (also called the nuclear- or Schatten $l_1$-norm). The trace-norm ball is the convex hull of the rank-1 matrices, which is also compact. The averaged multi-class objectives $f_i(\pmb{w})$ are
%\begin{eqnarray}
%\label{eq:fw_multiclass_objective_function}
$f_i(\pmb{w}) = \sum\nolimits_{k=1}^K \pmb{y}_k^{(i)} \log(p_k^{(i)})$ 
%\nonumber
%\end{eqnarray}
with
%\begin{eqnarray}
$p_k^{(i)} = {\exp({\pmb{w}\pmb{x}_i})}\slash{\sum\nolimits_{j = 1}^N \exp({\pmb{w}\pmb{x}_j})}\;.$ %\nonumber
%\end{eqnarray}
The individual gradients are
%\begin{eqnarray}
 $\nabla f_i(\pmb{w}) = (p_k^{(i)} - \pmb{y}_k^{(i)}) \pmb{x}^{(i)}\;.$
%\end{eqnarray}

Since Schatten-norms are invariant under orthogonal transformations, we can employ the singular value decomposition (SVD) to minimize the induced linear subproblems. Therefore the main computational cost of a single FWLayer on a Schatten-norm domain remains the computation of the SVD of $\nabla F(\pmb{w}_t)$, which is in $O(\min\{mn^2 , m^2n\})$~\cite{fw_jaggi13}.
For bounded trace-norm, however, the subproblems can be solved by a single approximate eigenvector computation instead of a complete SVD, which is much more efficiently, especially if the matrix dimensions are large and the optimal solution is low-rank \cite{fw_linear_convergence_trace_norm}. This gives Frank-Wolfe a significant computational advantage over projected and proximal gradient descent approaches. The vectors $\pmb{u}_1$ and $\pmb{v}_1$ can be efficient computed via power iteration~\cite{zheng17}.
This results in a rank-1 solution of Eq.~\eqref{eq:fw_multiclass_objective_function}, which can be written as $-\mu \pmb{u}_1\pmb{v}_1^T\;,$ where $\pmb{u}_1$ and $\pmb{v}_1$ are the unit left and right top singular vectors of the gradient matrix $\nabla F(\pmb{w}_t)$: %
%\begin{eqnarray}
 $\pmb{w}_{t+1} = (1-\gamma_t)\pmb{w}_t - \gamma_t\mu \pmb{u}_1\pmb{v}_1^T\;.$
%\end{eqnarray}

The Frank-Wolfe algorithm and corresponding FWNet for the trace-norm domain are shown in Fig.~\ref{fig:networks_fw_trace_with_power_method}. Here, lines 4-11 instantiate the general Frank-Wolfe algorithm in Alg.~\ref{alg:fw} with $k$ power iterations to compute the top singular vectors $\pmb{u}_1$ and $\pmb{v}_1$ of $\nabla F(\pmb{w}_t)$. \citeauthor{zheng17}~(\citeyear{zheng17}) showed that a small number of power iterations $K(t) = \mathcal{O}(\log t)$ is sufficient to ensure a sublinear convergence in expectation and if the number of power iterations are constant (i.e. $K(t) = k$ for all $t$) the Frank-Wolfe algorithm converges in expectation to a neighborhood of the optimal solution whose size decreases with $k$. In any case, the power iteration can naturally be implemented within FWNets using an RNN as summarized in Fig.~\ref{fig:networks_fw_trace_with_power_method}. Everything else remains conceptually the same and, in turn, we may even realize L2LC over low-rank matrices following similar arguments as for the unit simplex. 

\section{Experimental evidence}
\label{experiments}
Our intention here is to evaluate neural conditional gradients by investigating the following questions:
(\textbf{Q1}) Can FWNets compete with popular, non-neural
gradient descent approaches such as ADAM~\cite{kingmaB14}?
(\textbf{Q2}) Can we train CSVMs end-to-end using FWNets?
(\textbf{Q3}) Can L2LC be faster than L2LG?
(\textbf{Q4}) Do neural rank-1 softmax classifiers perform and scale well?

%\begin{figure}[t]
%	\begin{center}
%		\includegraphics[width=0.6\columnwidth]{images/cifar-2-svm/loss_zoomed}
%		\caption{Loss while training of FWNet, GD, L2LC, and L2L on different Cifar-10 subsets with two classes (Cifar-2). The %optimizee (L2LC and L2L) is trained on two fixed classes, but used to optimize all combinations of classes for the binary SVM %and therefore transferring to a completely novel dataset.\label{fig:cifar-2_loss-all}}
%	\end{center}
%     %\vspace{-5mm}
%\end{figure}

\begin{figure}[t]
	\begin{center}
    \begin{subfigure}[\scriptsize FWNets 500 iters.]{\label{fig:circles_hyperplane_3}
				\includegraphics[width=0.2\columnwidth]{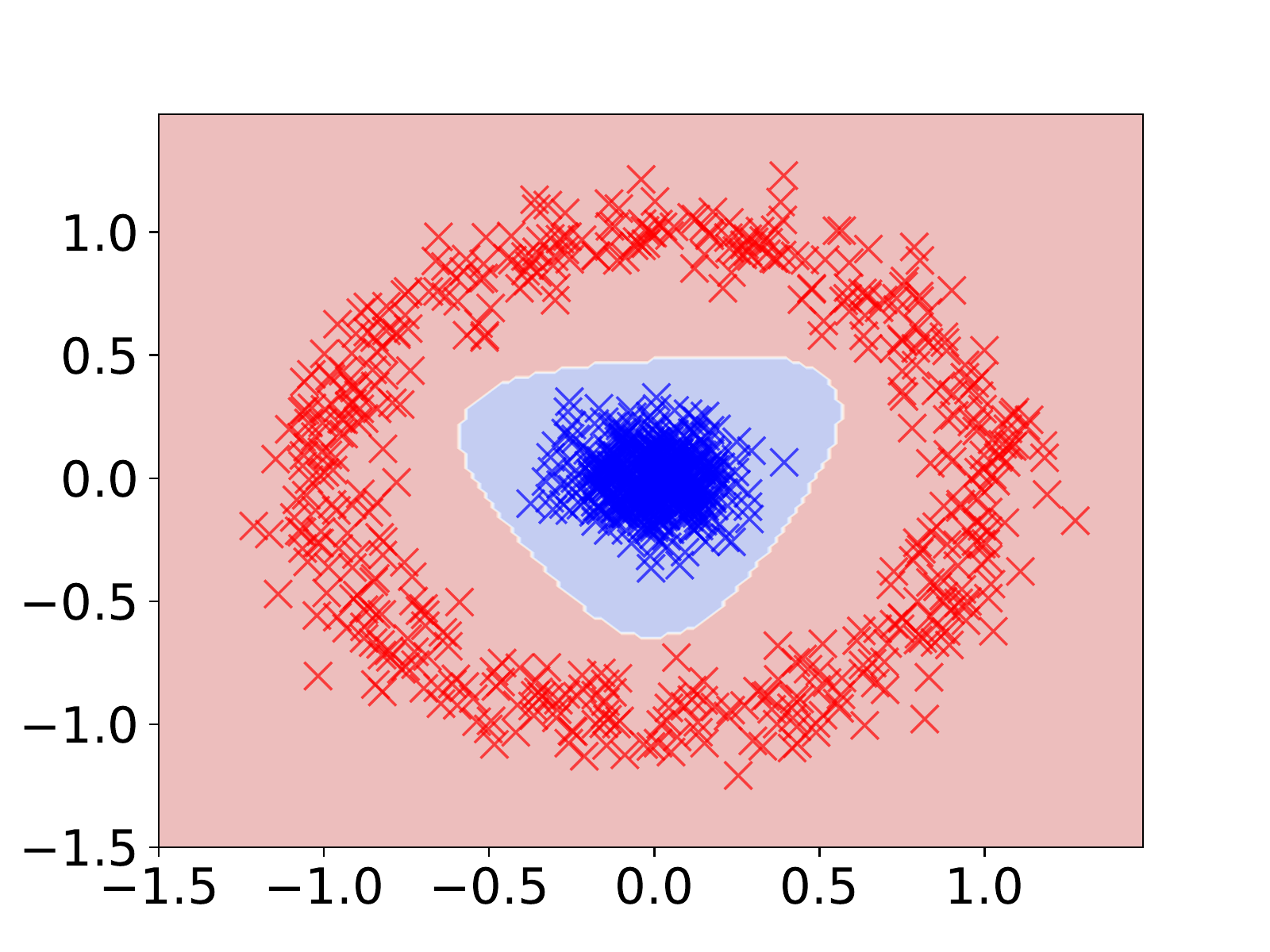}}
		\end{subfigure}
		\begin{subfigure}[\scriptsize L2LC$_\gamma$ 200 iters.]{\label{fig:circles_hyperplane_1}
				\includegraphics[width=0.2\columnwidth]{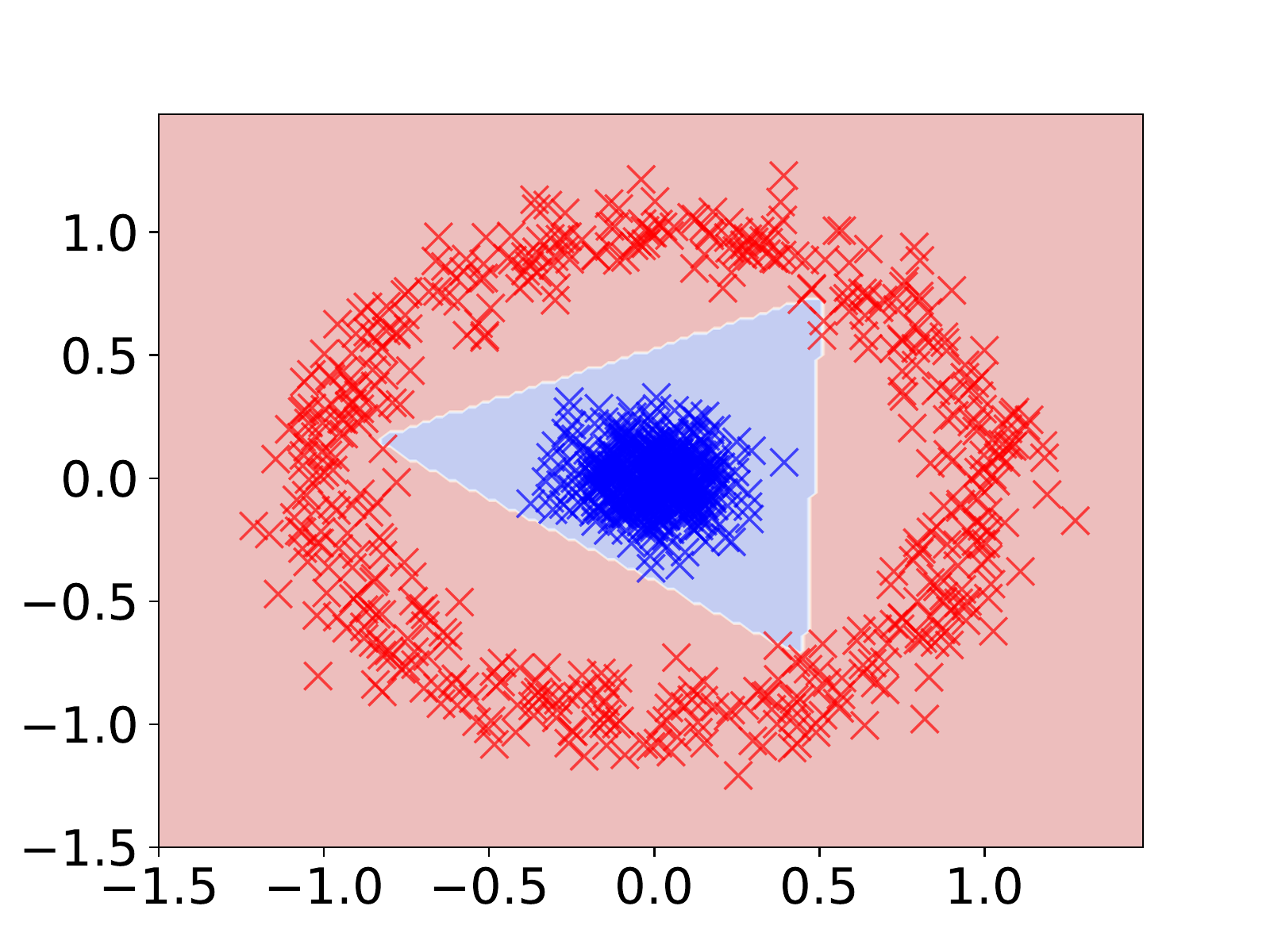}}
		\end{subfigure}
        	\begin{subfigure}[\scriptsize L2LC$_\nabla$ 20 iters.]{\label{fig:circles_hyperplane_2}
				\includegraphics[width=0.2\columnwidth]{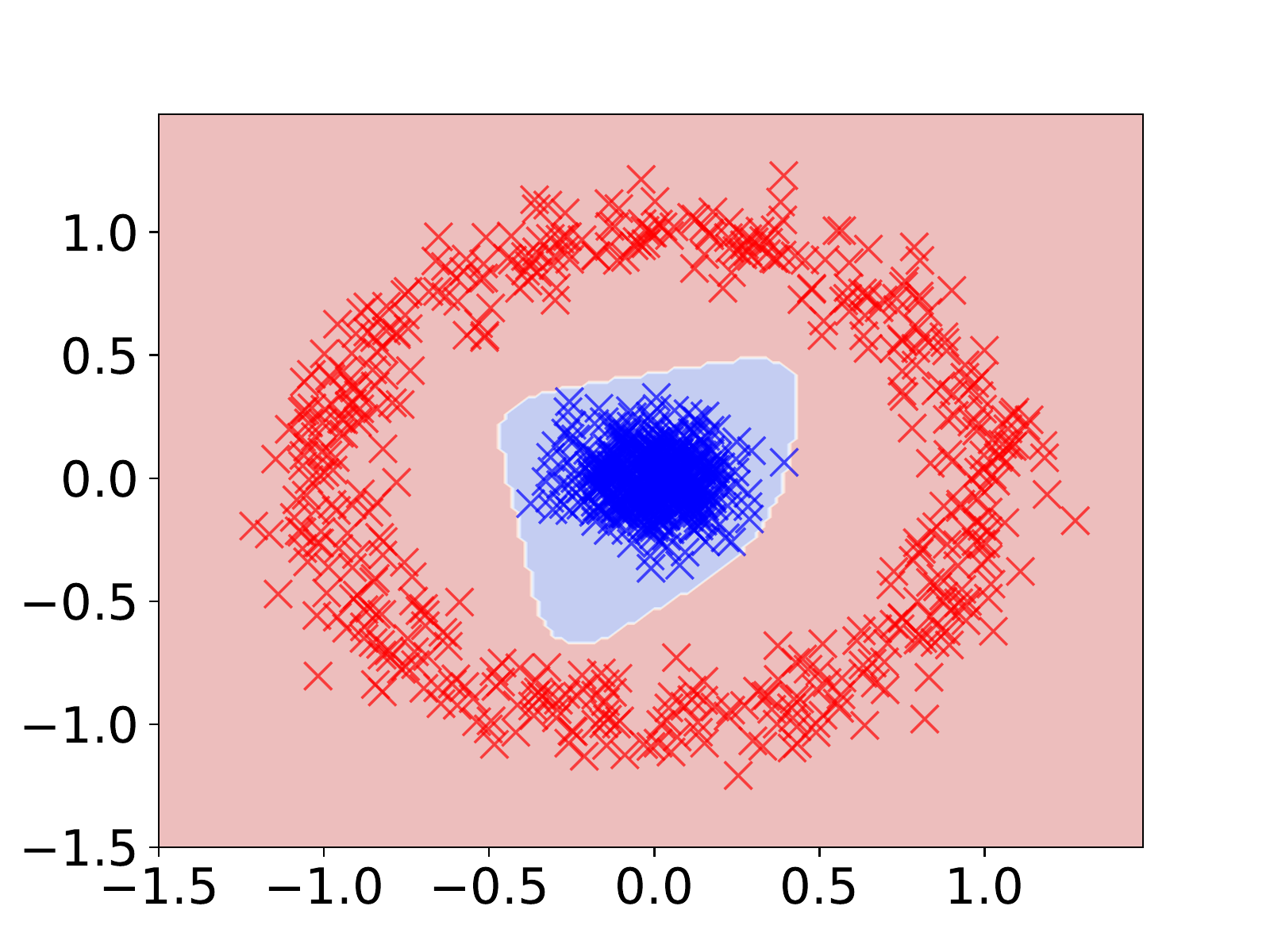}}
		\end{subfigure}
 \caption{Hyperplanes achieving the same losses are found at different iterations.
        The hyperparameters used were $\beta=1.0$, $C=1.0$, $\gamma_{FW}=0.01$, $\gamma_{ADAM}=0.01$ \label{fig:circles_hyperplanes}}        
	\end{center}
\end{figure}

To this end, we implemented FWNets, neural SVMs, neural softmax classifiers, and L2LC using the TensorFlow API
%\footnote{\url{https://www.tensorflow.org/versions/r1.3}} 
version 1.3 and the L2L implementation of \cite{l2l}. All experiments were ran on a Linux Machine with a NVIDIA GeForce GTX 1080 Ti with 11 GB memory and a AMD Ryzen Threadripper 1950X CPU with 16 physical cores having 32 threads in total. We considered several datasets. For comparing FWNets with classical, non-neural gradient optimization, we used both the synthetic datasets of ``concentric circles'' (Fig. \ref{fig:circles_hyperplanes}) as well as the real-world datasets MNIST~\cite{mnist} containing images of handwritten digits and Cifar-10 respectively Cifar-100~\cite{cifar} containing images of different animals and vehicles. For the L2L experiments, we split the data into three disjoint sets. One split was used to train the optimizee, one for training the optimizer, and the final one to test the corresponding learned model.
The neural SVMs are compared to ADAM gradient optimizers based on (1) reparameterization and (2) Lagrange multipliers to deal with the ``sum to one'' constraint. The objective of the Langrangian approach reads 
\begin{equation}
\label{lagrange_obj_func}
f(\pmb{\alpha}) = 0.5\pmb{\alpha}^T\pmb{K}\pmb{\alpha} - \lambda \sum\nolimits^{m}_{i = 1} \alpha_i\ \text{, s.t. } \alpha_i \geq 0.
\end{equation}
Furthermore we evaluate the performance of learning to learn the optimizer for solving the given problem. For that we used an LSTM as optimzee. More precisley, following \cite{l2l}, we introduced an additional LSTM to optimize the step-size respectively two different LSTMs when optimizing the the fully connected and convolutional layers. In all experiments we used two-layer LSTMs with 20 hidden units in each layer, 
aiming at minimizing \eqref{eq:dual_svm} respectively \eqref{lagrange_obj_func} using truncated backpropgation through time and early stopping in order to avoid overfitting. 
 
%\begin{figure}
%	\begin{center}
%		\includegraphics[width=0.6\columnwidth]{images/circles/normal_setup/loss}
%		\caption{Loss while learning ``concentric circles'' using $\beta=1.0$, $C=1.0$, $\gamma_{FW}=0.01$, %$\gamma_{ADAM}=0.01$. L2LC$_\gamma$ (green line) optimizes the step-size and L2LC$_\nabla$ (blue line) the kernel. %\label{fig:circles_loss}}
%	\end{center}
%    %\vspace{-1mm}
%\end{figure}

\begin{figure}[t]
\begin{center}
\begin{minipage}[t]{\dimexpr.5\textwidth-0.5em}
\begin{center}
		\includegraphics[width=1.0\columnwidth]{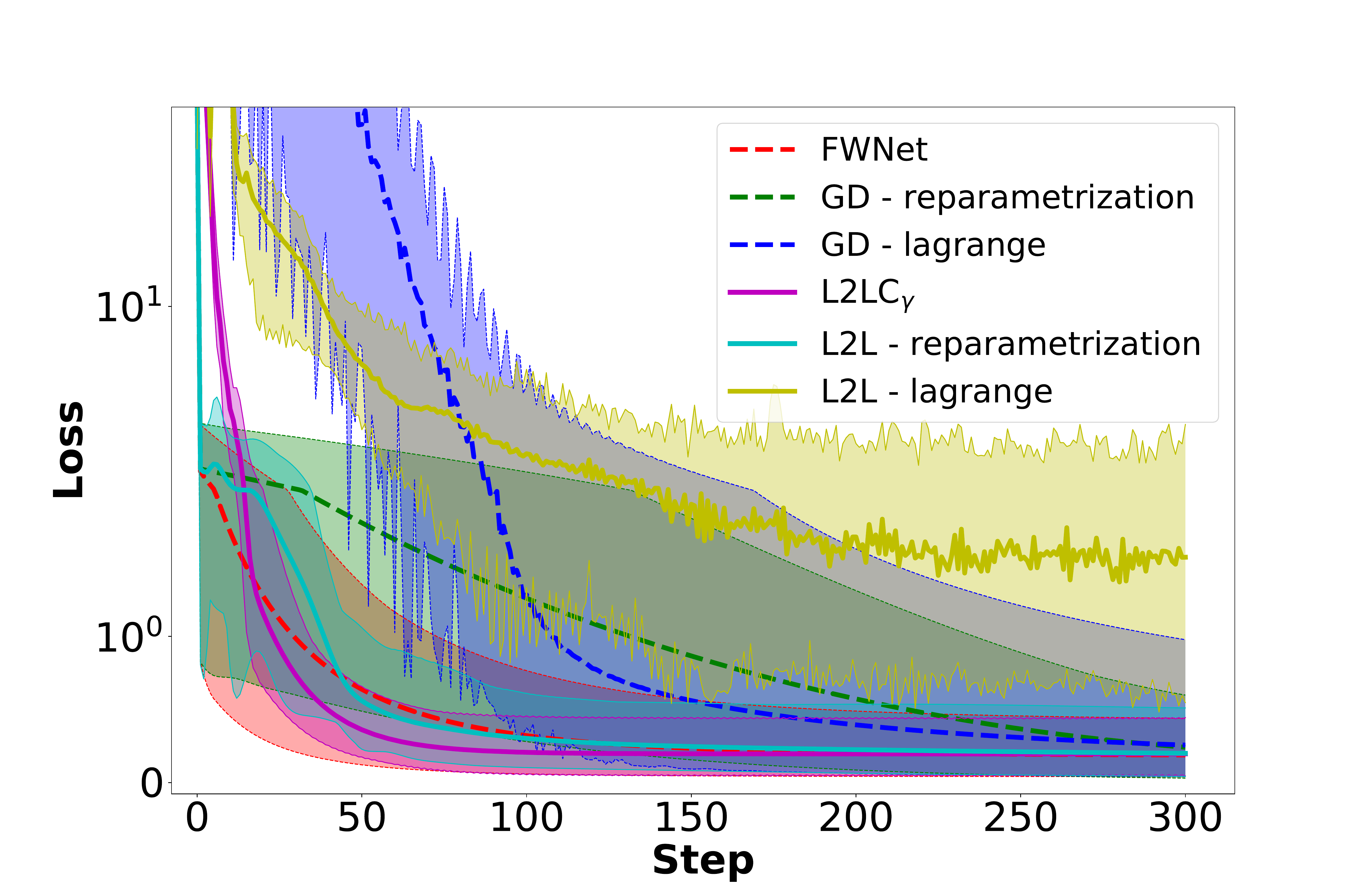}
		\caption{Loss while training of FWNet, GD, L2LC, and L2L on different Cifar-10 subsets with two classes (Cifar-2). The optimizee (L2LC and L2L) is trained on two fixed classes, but used to optimize all combinations of classes for the binary SVM and therefore transferring to a completely novel dataset.\label{fig:cifar-2_loss-all}}
        \end{center}
\end{minipage}\hfill
\begin{minipage}[t]{\dimexpr.5\textwidth-0.5em}
\begin{center}
		\includegraphics[width=1.0\columnwidth]{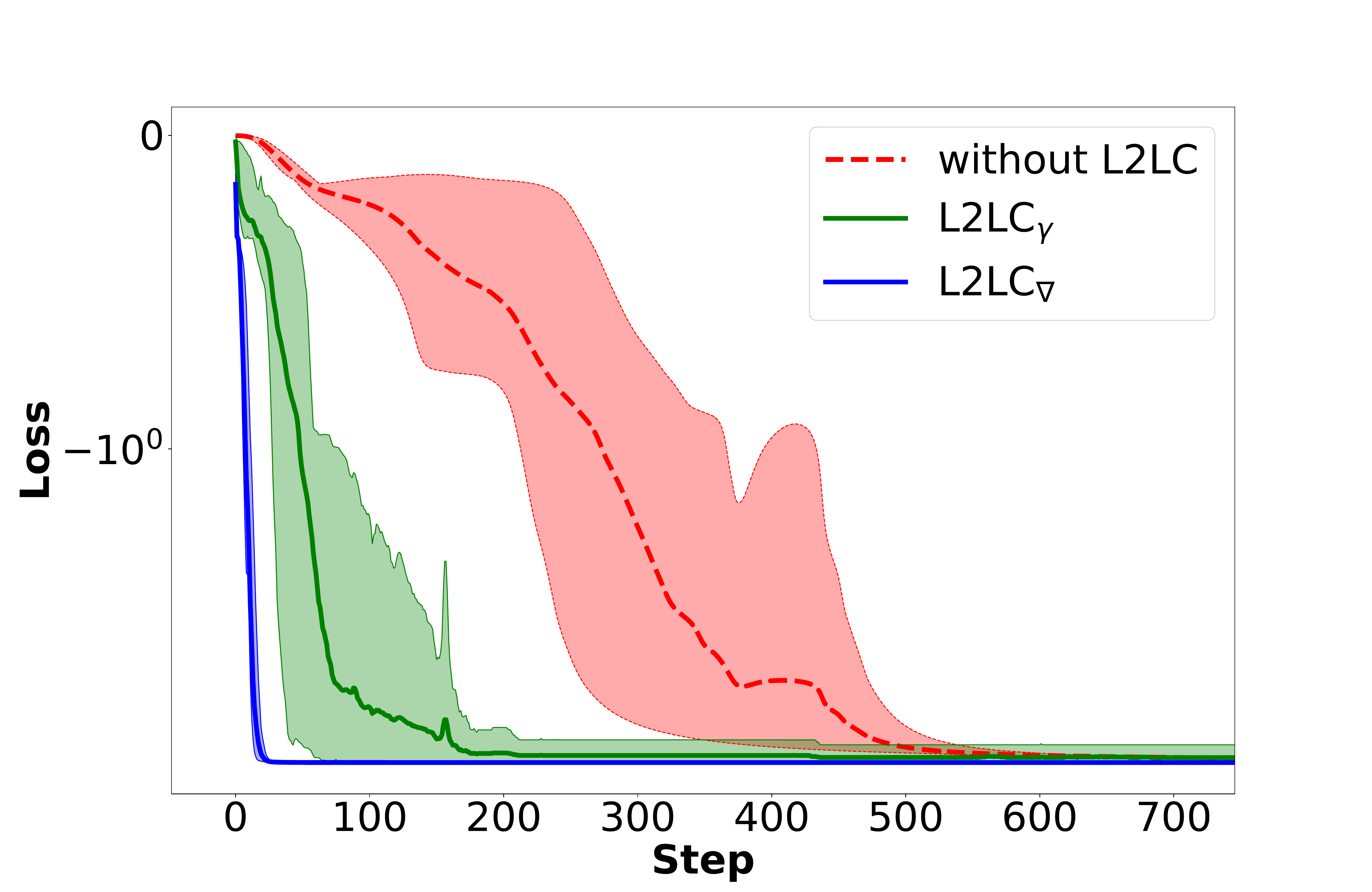}
		\caption{Loss while learning ``concentric circles'' using $\beta=1.0$, $C=1.0$, $\gamma_{FW}=0.01$, $\gamma_{ADAM}=0.01$. L2LC$_\gamma$ (green line) optimizes the step-size and L2LC$_\nabla$ (blue line) the kernel. \label{fig:circles_loss}}
        \end{center}
\end{minipage}
\end{center}
\end{figure}

%\subsection
{\bf Few-Shot Neural SVMs (Q1, Q3).}
In our first experiment we considered  classes 1 and 2, denoted as Cifar-2, from the Cifar-10 dataset. We extracted their features from an inception-network and used them for training the base models (\textbf{Q1}) using a linear kernel. Additionally we train an optimizee for FW (\textbf{Q3}). A random search set $\beta = 10$. Fig.~\ref{fig:cifar-2_loss-all} summarizes the results. The optimizee is trained on the classes 1 and 2, but then used to optimize neural SVMs on all pairwise combinations (1-3,1-4,...,2-3,...) of classes from Cifar-10 and therefore transferring to a completely novel dataset. As one can see, FWNets and L2LC outperformed the other baselines. FWNets with an hand-design, adaptive stepsize can be slightly faster than L2LC$_\gamma$, but the LSTM learns to control FW in a similar way and shows a much smaller variance. This answers (\textbf{Q1}, \textbf{Q3}) affirmatively.

%\subsection
{\bf Deep SVMs (Q2, Q3).}
Next we considered training deep SVMs on the ``concentric circles'' dataset, i.e., we placed an FWNet as a last layer of a neural network, trained in an end-to-end (\textbf{Q1}) as well as in a learning to learn fashion (\textbf{Q4}). The neural network we used as kernel contained three layers. The first and second layer are fully-connected layer with four and two neurons each, trained using ADAM. 
Fig.~\ref{fig:circles_loss} summarizes the results. As one can see, the learned optimizers outperformed the hand-coded ones. Moreover, as Fig.~\ref{fig:circles_hyperplanes} illustrates,
a learned optimizer may find smoother hyperplanes achieving the same loss in less many iterations, when also adapting the kernel: just using FWNets takes 500 iterations; when the LSTM controls the step size, it takes 200 iterations; when also adapting the kernel, it just takes 20 iterations and the hyperplane is considerably smoother. 
This answers (\textbf{Q2}, \textbf{Q3}) affirmatively.

\begin{figure}[t]
	\begin{center}
		\includegraphics[width=0.5\columnwidth]{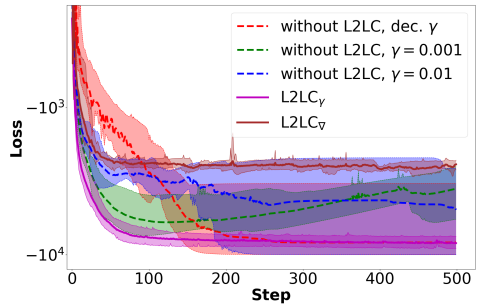}
		   \caption{Loss while training on the Cifar-10 dataset using the two classes \textit{manmade} and \textit{natural}. \label{fig:cifar_manmade_vs_natural}}
	\end{center}
    \vspace{-2mm}
\end{figure}

To investigated deep SVMs further, we also considered Cifar-10. 
We split the labels of the dataset in two different classes, namely \textit{natural} and \textit{manmade}. The class \textit{natural} contains the classes %cifar10-classes 
\textit{bird}, \textit{cat}, \textit{deer}, \textit{dog}, \textit{frog} and \textit{horse}, and the \textit{manmade} class the classes \textit{airplane}, \textit{automobile}, \textit{ship} and \textit{truck}. As kernel we used a neural network with both convolutional and fully-connected layers: three convolutional layers with max pooling followed by a fully-connected layer with 32 hidden units; all non-linearities were ReLU activations with batch normalization. The final layer is a FWNet simulating to train an SVM, and the rest of the network was trained using ADAM. Fig.~\ref{fig:cifar_manmade_vs_natural} summarize the results.
As one can see, the stepsize-learned conditional gradient L2LC$_\gamma$ outperforms hand-coded optimizers even with adaptive stepsize; requiring less than half of the iterations to converge. Training also the kernel is harder as it is a non-convex problem; exploring this further is an interesting avenue for future work. In any case, the results answer (\textbf{Q2}, \textbf{Q3}) affirmatively.

\begin{figure}[t]
	\begin{center}
		\includegraphics[width=0.37\columnwidth]{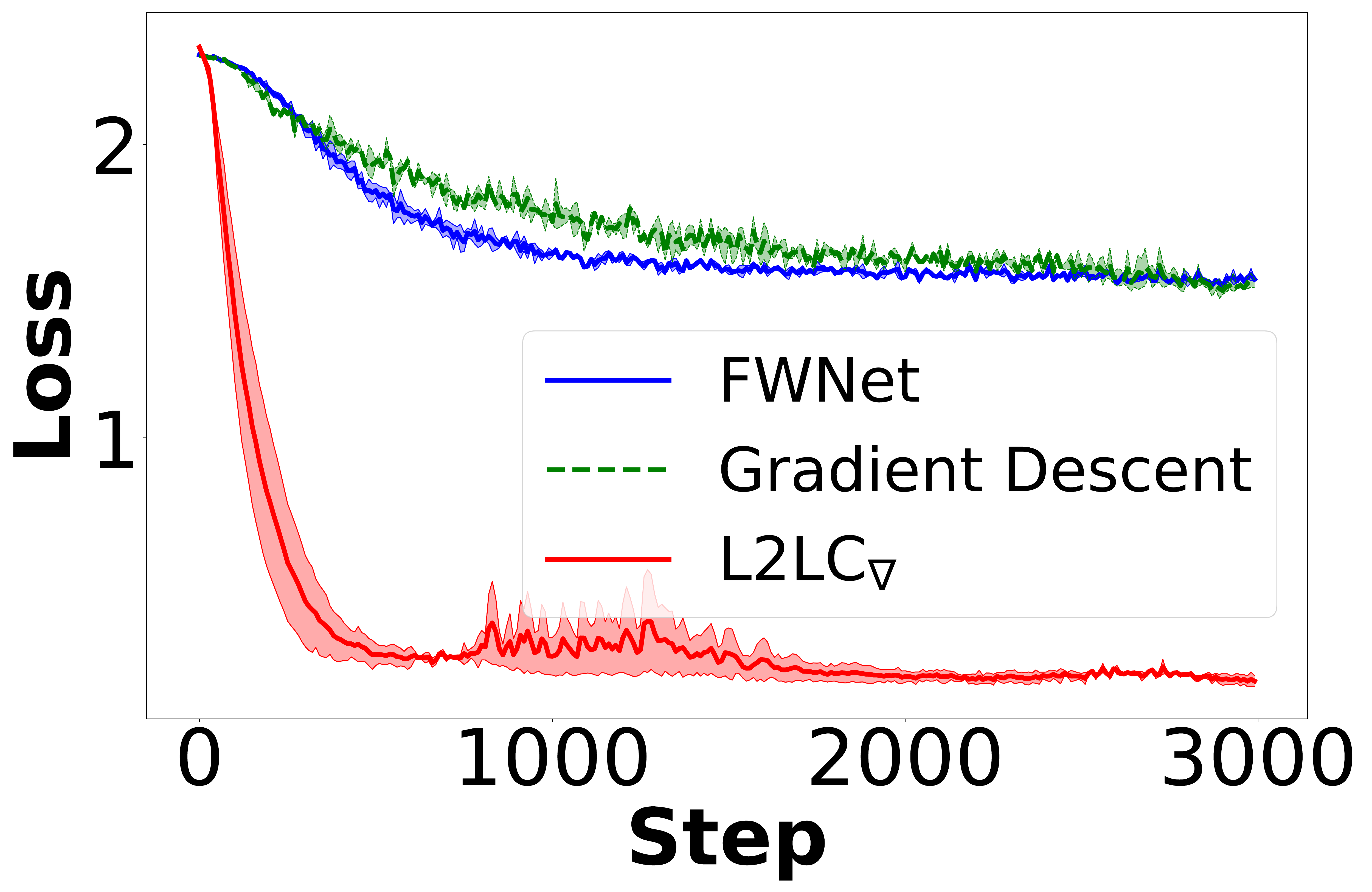}
		\includegraphics[width=0.37\columnwidth]{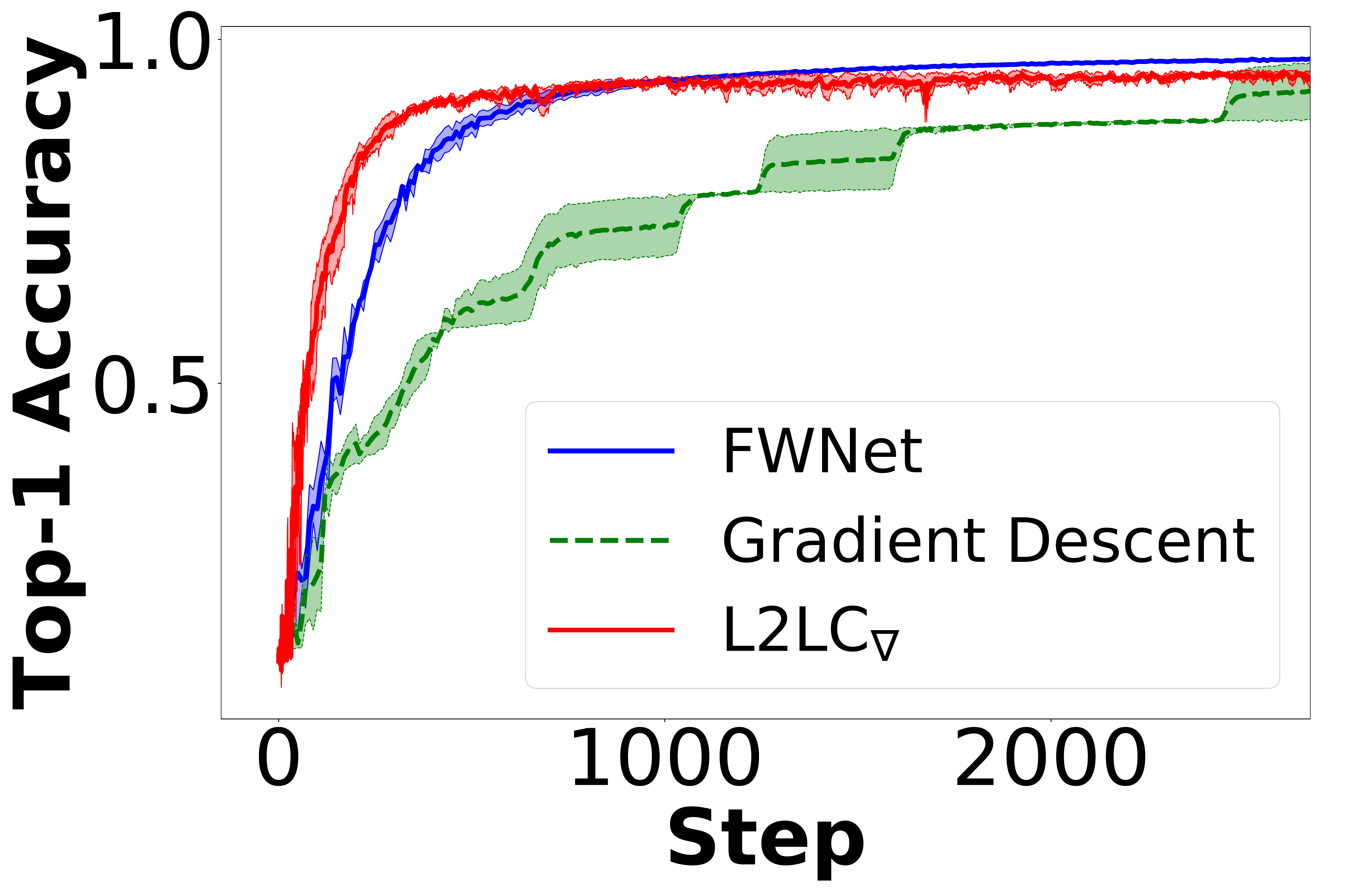}
        \caption{(Left) Learning curves of FWNets- (blue line), L2LC$_\nabla$ (red-line) and ADAM-based (green line) softmax classifiers on the MNIST dataset. (Right) Top-1 accuracy on test-set on MNIST.
        \label{fig:trace_norm_mnist_cnn_loss}}
	\end{center}
    \vspace{-2mm}
\end{figure}

%\subsection
{\bf Sparse Neural Softmax Classifiers (Q1, Q3, Q4).}
Finally, we investigated FWNets for training deep softmax classifiers on MNIST. We used a simple CNN with two convolutional layer and one fully-connected consisting of 16 neurons followed by a fully-connected softmax layer. For both FWnets and ADAM-based optimizers we used the same constant step size of $\gamma=0.001$. The FWNets unrolled the power iteration networks for five steps in order to compute the left and right top singular vectors of the gradient matrix $\nabla F(\pmb{w}_t)$. The step size was set $\mu= 50$. Additionaly we train an optimizee (L2LC$_\nabla$) for FW. Therefore we split the training-set of MNIST in two disjoint sets with the same size. The results are summarized in Fig.~\ref{fig:trace_norm_mnist_cnn_loss}.
As one can see, the sparse FWNet and L2LC$_\nabla$ outperformed ADAM, both in terms of convergence and predictive performance; the same top-1 accuracy in less than third of the iterations. Furthermore the trained classifier optimized with the optimizee results in a much higher confidence of the predictions, as one can see from the behavior of the Loss-function.
This answers (\textbf{Q1}, \textbf{Q2}, \textbf{Q4}) affirmatively.
To investigate this further, we also considered wider and deeper CNNs on MNIST and also on Cifar-10 and Cifar-100. The more dense the network became, the better the ADAM performed. This validates our assumption of low-rank solutions: if the low-rank assumption does not hold or is not required, there is no point in estimating a sparse model using the trace-norm constraint~\cite{fw_linear_convergence_trace_norm}.
%\begin{figure}[t]
%	\begin{center}
%		\includegraphics[width=0.4\columnwidth]{images/trace_norm/mnist_cnn/loss}
%		\includegraphics[width=0.4\columnwidth]{images/trace_norm/mnist_cnn/acc_test}
%        \caption{(Left) Learning curves of FWNets- (blue line) and ADAM-based (green line) softmax classifiers on the MNIST %dataset. (Right) Top-1 accuracy on test-set on MNIST.
%        \label{fig:trace_norm_mnist_cnn_loss}}
%	\end{center}
%    \vspace{-2mm}
%\end{figure}

\section{Conclusion}
We have introduced the {\it learning to learn by conditional gradients} (L2LC) framework based on Frank-Wolfe Networks (FWNets). This enables one to train sparse convex optimizers that are specialized to particular classes of problems. We illustrated this for training SVMs and sparse softmax classifiers. Our experimental results confirm that learned conditional gradients compare favorably against state-of-the-art optimization methods used in deep learning. 

There are several interesting avenues for future work. 
One should develop FWNets for other ML tasks such as graph classification~\cite{kerstingMGG14} and Bayesian Quadrature~\cite{briolOGO15} as well as for other FW approaches ~\cite{fw_jaggi13}. One may also adapt the Power Iteration in an end-to-end fashion~\cite{duvenaudMABHAA15}. Finally, 
hierarchical RNNs~\cite{wichrowskaMHCDF17} have the potential to speed up {\it learning to learn by conditional gradients}.

{\bf Acknowledgments: }This work was supported by the Federal Ministry of Food, Agriculture and Consumer Protection (BMELV) based on a decision of the German Federal Office for Agriculture and Food (BLE); grant nr. ``2818204715''.

\bibliography{paper}
\bibliographystyle{icml2018}

%#########################################################################################################################################

%\end{document}

%#########################################################################################################################################

\end{document}